\documentclass[11pt]{article}
\usepackage{verbatim}
\usepackage[final]{acl}
\usepackage[toc,page]{appendix}
\usepackage{times}
\usepackage{latexsym}
\usepackage[T1]{fontenc}
\usepackage[utf8]{inputenc}
\usepackage{microtype}
\usepackage{graphicx}
\usepackage{multirow}
\usepackage{amsmath}
\usepackage{amssymb}
\usepackage{xcolor}
\usepackage{booktabs}
\usepackage{tabularx}
\usepackage{array}
\usepackage{float}
\usepackage{longtable}
\usepackage{afterpage}
\usepackage{natbib}
\usepackage{lipsum}
\usepackage{inconsolata}
\usepackage[most]{tcolorbox}

\newtcolorbox{promptsystem}[1]{%
    enhanced,
    breakable,
    width=\textwidth,
    colback=white,
    colframe=black!55,
    colbacktitle=black!8,
    coltitle=black,
    boxrule=0.5pt,
    arc=0.8mm,
    left=2mm,
    right=2mm,
    top=1.5mm,
    bottom=1.5mm,
    before skip=5pt,
    after skip=5pt,
    fontupper=\footnotesize,
    fonttitle=\small\bfseries,
    title={#1}
}

\newtcolorbox{agentprompt}[1]{%
    enhanced,
    breakable,
    colback=black!2,
    colframe=black!25,
    colbacktitle=black!6,
    coltitle=black,
    boxrule=0.35pt,
    arc=0.5mm,
    left=2mm,
    right=2mm,
    top=1mm,
    bottom=1mm,
    before skip=2pt,
    after skip=2pt,
    fontupper=\footnotesize,
    fonttitle=\footnotesize\bfseries,
    title={#1}
}

\title{SIREN: Situation-aware Index for Response and Emergency ActioNability in Crisis Translation}

\author{
  \textbf{Maria Carmen Staiano\textsuperscript{1}},
  \textbf{Lifeng Han\textsuperscript{2}},
  \textbf{Johanna Monti\textsuperscript{3}},
  \textbf{Francesca Chiusaroli\textsuperscript{1}}
  \\[2mm]
  \textsuperscript{1}University of Macerata 
  \textsuperscript{2}LUMC \& Leiden University 
  \textsuperscript{3}University of Naples ``L'Orientale''\quad
  \\[1mm]
  \texttt{m.staiano@unimc.it},
  \texttt{l.han@lumc.nl},
  \texttt{jmonti@unior.it},
  \texttt{f.chiusaroli@unimc.it}
}

\begin{document}
\maketitle

\begin{abstract}
Machine translation (MT) is increasingly utilized in multilingual crisis communication, yet conventional evaluation does not directly assess whether translated emergency messages preserve their actionability. We introduce SIREN (Situation-aware Index for Response and Emergency ActioNability), a human evaluation framework assessing whether translation choices compromise protective action, directive force, and urgency of the source text. We evaluate SIREN in two complementary settings. At the error level, we apply it to HOPE-annotated Italian--English translations produced by ChatGPT-4o and Google Translate: 31.63\% and 48.73\% of errors, respectively, compromise actionability, predominantly by affecting the protective action. At the message level, we evaluate English--French translations produced by TranslateGemma-4B and Google Translate, with actionability preserved in 85.34\% and 92.67\% of translations, respectively. TranslateGemma-4B failures predominantly weaken directive force, whereas Google Translate failures more often compromise protective actions and receive Major impact ratings. These findings show that translation systems can differ not only in how often actionability is lost, but also in the nature and functional impact of these failures. Alongside SIREN, we contribute an English--French crisis-translation resource with human actionability annotations and post-edited reference translations, supporting further research on function-oriented evaluation for crisis communication.\footnote{accepted at
The First Workshop on Language Technologies for Crisis Preparedness and Response (LT4CPR) at AACL-IJCNLP 2026 \url{https://2026.aaclnet.org/}
}Our experimental data is available at \url{https://github.com/mcstaiano/SIREN}.
\end{abstract}

\section{Introduction}
\label{sec:intro}
Artificial intelligence (AI) is shaping how warnings are delivered, making them faster and easier to access. Current AI systems can translate  emergency alerts into multiple languages, tailor messages to different audiences and contexts, and adapt dissemination strategies in real time across communication channels \citep{itu2026ai}. International frameworks such as the Sendai Framework for Disaster Risk Reduction \cite{unisdr2015sendai} and the United Nations Sustainable Development Goals (SDGs) \cite{un2015transforming} call for multilingual early warning systems that can reach entire populations, including linguistic minorities, remote communities, displaced populations and tourists who do not speak the local language. 

Meeting this multilingual communication need at scale increasingly depends on the use of automatic translation. Yet, when disseminating crisis messages, linguistic adequacy on its own does not ensure effective reception by target users. Emergency warnings may be multimodal, time-sensitive or anchored to specific cultural contexts \cite{o2020crisis}. Their evaluation should therefore extend beyond the linguistic surface to consider the \textbf{actionability} of the translated message, that is, \textit{whether it enables recipients to identify and perform the intended protective action} \cite{sengupta2024transcreation}.

This perspective resonates with Nida's functional equivalence theory, according to which ``translation should also aim to produce the same effect on the reader as the original text did'' \cite{nida1964toward}.
In crisis communication, it invites us to reconsider how Machine Translation (MT) quality is assessed. The question is not only whether a translation preserves the meaning of the source, but also \textit{whether it supports the behavioural response that the message is intended to elicit}. This motivates the central question of our study: \textbf{Does a translated emergency message preserve the actionability of its source?}

To address this question, we introduce \textbf{SIREN} (\textbf{S}ituation-aware \textbf{I}ndex for \textbf{R}esponse and \textbf{E}mergency Actio\textbf{N}ability), a human evaluation framework designed to assess \emph{actionability preservation} in emergency-message translation. We define actionability preservation as the extent to which a translated message retains the capacity of the source to enable recipients to identify and perform the intended protective action. Rather than assessing general translation quality, SIREN focuses on whether translation choices may affect the protective behaviour expected from recipients. It thus complements semantic and error-based MT evaluation with a functional, pragmatically-oriented perspective.

\section{Related Work}
\label{sec:related_work}

\paragraph{MT evaluation in the Crisis Domain.}
Current approaches to MT evaluation in the crisis domain revolve around two main paradigms: (1) an output-focused approach, which assesses translation quality through textual and linguistic properties of the translated output, and (2) a reception-oriented approach, which focuses on how end users understand, interpret, and act upon translated information \cite{hao2026quality}. 
Within the output-focused paradigm, translations can be evaluated using automatic metrics such as lexicon-based BLEU \cite{papineni-etal-2002-bleu}, METEOR \cite{lavie-agarwal-2007-meteor}, LEPOR \cite{han-etal-2012-lepor}, chrF \cite{popovic-2015-chrf}, embedding-space COMET \cite{rei_comet-22_2022} and BERTScore \cite{BERTScoreDBLP:conf/iclr/ZhangKWWA20}, as well as human evaluation frameworks such as Multidimensional Quality Metrics (MQM) \cite{lommel_tutorial_2018,lommel2024multi}.

Previous research adopting this perspective in crisis MT has primarily examined translation quality through automatic metrics \cite{anastasopoulos-etal-2020-tico, lankford2024leveraging}, human error analysis, or a combination of the two \cite{castaldo2026translating, robinson-2025-take, staiano2025italert}. For instance, Staiano et al.~\shortcite{staiano2025italert} investigated the performance of neural machine translation and LLM-powered automatic translation using ITALERT, a parallel Italian--English corpus covering eight crisis scenarios. Combining automatic metrics with human error analysis based on an MQM-derived framework, the study highlighted the limitations of relying exclusively on automatic evaluation scores. Nevertheless, Ticona and Anastasopoulos~\shortcite{Ticona_Anastasopoulos_2026} noted that research evaluating MT in the crisis domain using LLMs remains limited, underscoring the insufficient coverage of low-resource languages.

Severity-aware human evaluation further makes it possible to account for the potential consequences of translation errors that may not be reflected in aggregate automatic scores, thereby supporting more informed assessments of whether MT is suitable for deployment in crisis scenarios \cite{cadwell2019more}. 

Recent work has also broadened this perspective by considering MT \emph{usability} and \emph{readiness}, shifting attention from relative system performance towards whether translation quality is sufficient for practical use in crisis response \cite{bansal2026mt}. Other work has begun to examine the preservation of crisis-relevant communicative properties. Investigating MT alongside LLM-based \emph{urgency} assessment, Ticona and Anastasopoulos~\shortcite{Ticona_Anastasopoulos_2026} show that translation and prompting choices can systematically alter perceived urgency across languages. Altogether, these approaches reflect a growing concern with evaluating MT not only in terms of translation quality, but also in relation to its role in downstream crisis-response tasks, such as disaster situation assessment and automatic triage.

\paragraph{Actionability in Crisis Translation.}
Reception-oriented approaches offer a complementary perspective by shifting the focus from linguistic properties of the translated output to its pragmatic use by intended recipients. In crisis communication, this perspective has been
associated with dimensions such as \emph{timeliness}, \emph{understandability},
\emph{trustworthiness}, and \emph{actionability} \cite{hao2026quality}. Among these, actionability is particularly relevant to emergency messages containing protective instructions, as their communicative purpose is not only to convey
information but also to guide recipients towards an appropriate response.

Actionability has already been investigated empirically in the reception of machine-translated emergency messages. \newcite{hajek2024understanding} examined the viability of MT for emergency communication from a reception-oriented perspective, comparing raw MT, human post-edited MT, and MT of pre-edited source texts with speakers of Spanish, Greek, Chinese and Dari. Their study investigated how recipients understood and responded to machine-translated emergency information, including whether translation inaccuracies affected their decisions about the actions to take. 

Building on this line of research, we approach actionability from a
translation-evaluation perspective. While Hajek et al.~\shortcite{hajek2024understanding} investigate actionability through
recipients' responses to different MT workflows, we shift the unit of analysis
to the source--translation relationship, asking whether, and to what extent,
the behavioural function encoded in the source emergency message is preserved
in translation. We further broaden the technological scope
of previous work by investigating actionability preservation across both
commercial MT systems and open-weight LLMs.


\section{Methodology}
\label{sec:methodology}

\begin{figure}[t]
    \centering
    \includegraphics[width=.49\textwidth]{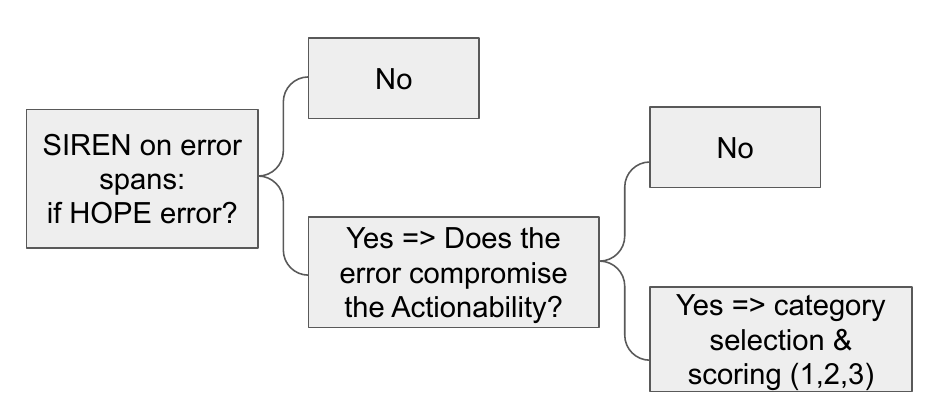}
    \caption{SIREN on Error Span Level.}
    \label{fig:siren-error-level}
\end{figure}

\begin{figure}[t]
    \centering
    \includegraphics[width=.49\textwidth]{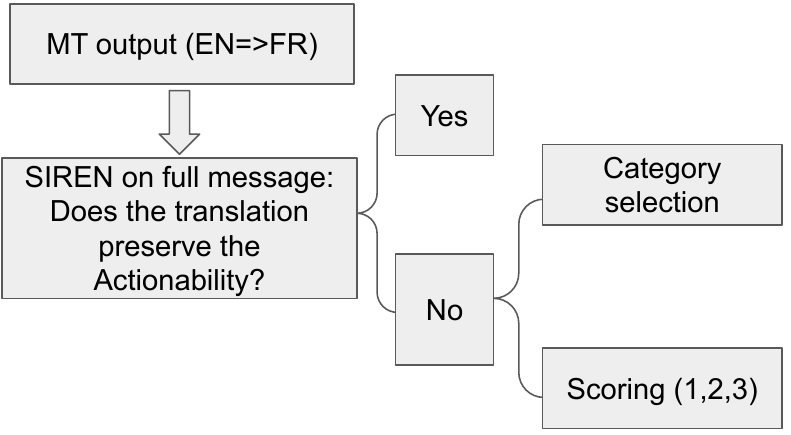}
    \caption{SIREN on Full Message Level.}
    \label{fig:siren-full-message-level}
\end{figure}

\paragraph{Overview.}
SIREN is applied in two variants that differ in their unit of analysis: the error-span level (SIREN-span; Figure~\ref{fig:siren-error-level}) and the full-message level (SIREN-msg; Figure~\ref{fig:siren-full-message-level}). In both, annotators first give a binary judgement (\textsc{Yes}/\textsc{No}). If actionability is compromised, they select one of four predefined rationale categories and assign an impact level on a three-point scale: \emph{Minor} (1), \emph{Medium} (2), or \emph{Major} (3).

\paragraph{Rationale categories.}
The four rationale categories are defined as follows:
 \textbf{Directive force weakened} indicates that the intended instruction becomes less direct, imperative, or insistent; \textbf{Directive force strengthened} indicates that it becomes more direct, imperative, or insistent than in the source; \textbf{Protective action compromised} indicates that the required action is changed, its scope is narrowed or broadened, or it becomes unclear or non-executable; and \textbf{Urgency marker omitted} indicates that an explicit or implicit urgency marker is omitted or attenuated, reducing the perceived immediacy of the required action.

The taxonomy was designed by drawing on complementary perspectives from pragmatics and applied frameworks for action-oriented communication.  First, the distinction between \emph{weakened and strengthened directive force} is informed by speech-act theory. Austin~\shortcite{austin1975things} distinguishes between locutionary, illocutionary and perlocutionary acts, while Searle~\shortcite{searle1976toward} further develops the theory, by classifying illocutionary acts into five categories, including directives, which aim to get the addressee to perform a specific action. This distinction allows SIREN to capture
translations in which the intended protective action remains identifiable, but the strength with which the directive is communicated changes. 

Second, the \emph{protective action compromised} category draws on the notion of actionability operationalised in the Patient Education Materials Assessment Tool (PEMAT) \citep{shoemaker2014development}, according to which materials are actionable when users of diverse backgrounds can identify what they can do on the basis of the information presented. Finally, \emph{urgency marker omitted} is motivated by the Common Alerting Protocol (CAP) \citep{cap2010}, which treats urgency as a distinct property of an emergency alert, indicating the timeframe in which a responsive action should be taken.

Taken together, these sources motivate three functional dimensions of actionability: the protective action itself, the directive force with which it is communicated, and the urgency associated with the alert message, which SIREN operationalises through the four rationale categories.

\paragraph{SIREN-span.}
At the error-span level, we build on error annotations produced with HOPE \citep{gladkoff2022hope}, an MQM-derived human evaluation framework that assigns translation errors a category and severity level. SIREN-span is applied as an additional annotation layer by asking, for each error identified by HOPE: \emph{Does the error compromise the actionability of the emergency message?} Since a single sentence can contain multiple errors, the number of error spans can be higher than the original number of sentences. If the answer is \textsc{Yes}, annotators select a rationale category and an impact level as described above.

\paragraph{SIREN-msg.}
For SIREN at the full-message level, annotators directly assess
the translated emergency message by answering the binary question:
\emph{Does the translation preserve the actionability of the source message?}
If actionability is not preserved, annotators follow the same decision
procedure as in SIREN-span: they select the rationale that best characterises
the functional change and assign an actionability impact level. 

The two settings provide complementary operationalisations of the same underlying construct. In the \textbf{error-level setting} (SIREN), annotators judge whether each individual translation error already compromises actionability. In the \textbf{message-level setting}, annotators judge whether the translated message as a whole preserves actionability, answer the same questions, and post-edit the message to restore it, producing a gold standard. The first operationalises actionability as a metric; the second provides an independent, holistic validation of that metric and a reusable, actionability-preserving resource.

\section{Experimental Work}
\label{sec:experimental work}

\subsection{Corpus}

We use data from the \textsc{ITALERT} corpus
\citep{staiano2025italert}, a bilingual Italian--English collection of authentic emergency-response messages drawn from the Italian Civil Protection campaign \emph{Io non rischio}\footnote{Source: ``Dipartimento della Protezione Civile-Presidenza del Consiglio dei Ministri''.}. The corpus comprises 440 segments across eight crisis subdomains: flooding, earthquake, forest fire, volcanic eruption, tsunami, industrial accident, nuclear risk, and dam failure. Each source segment is paired with a professional human reference translation, and the corpus is open-sourced for reusability\footnote{\url{https://github.com/mcstaiano/ITALERT}}.

\subsection{Study 1: Error-Level Actionability}
\label{sec:study1}

Study~1 evaluates actionability preservation at the level of individual translation errors. We use the Italian$\rightarrow$English translations from the \textsc{ITALERT} corpus produced by the two MT systems evaluated in the original study: ChatGPT-4o and Google Translate. First, we conduct pilot experiments with two annotators and calculate the inter-annotator agreement. Then, during the main annotation phase (rollout), the annotators divided the dataset and applied the SIREN protocol to 393 previously identified HOPE error instances: 196 from ChatGPT-4o and 197 from Google Translate.
Detailed annotator profiles are provided in Appendix~\ref{appendix:annotators}.

\subsubsection{Layer-I: Existing HOPE Annotations}
Each translation already carries error annotations previously produced with the HOPE framework \cite{gladkoff2022hope} by three professional linguists. Each error is assigned one of seven categories: Impact (IMP), Terminology (TRM), Ungrammatical (UGR), Mistranslation (MIS), Style (STL), Proofreading (PRF), Proper Name (PRN), along with a severity score (minor~1, medium~2, major~4, severe~8, critical~16). The original HOPE evaluation identified 611 errors for ChatGPT-4o and 728 for Google Translate. SIREN builds on these pre-existing annotations, without re-detecting translation errors or modifying their original HOPE severity.

\subsubsection{Layer-II: SIREN-span Annotation}
The second annotation layer applies SIREN-span to the pre-existing HOPE errors described in Layer~I. The individual HOPE error constitutes the unit of analysis: each annotated error span is treated as a separate SIREN instance, even when multiple errors occur within the same sentence. For each error, annotators answer a binary question:

\begin{quote}
\emph{Does this error compromise the actionability of the emergency message?}
\end{quote}

A \textsc{No} judgement indicates that, despite the presence of a translation
error, the intended protective action remains actionable. A \textsc{Yes}
judgement triggers two additional annotations following the SIREN-span
procedure described in Section~\ref{sec:methodology}: annotators identify the
functional rationale for the loss of actionability and assign an
actionability impact level (\textsc{Minor}, \textsc{Medium}, or
\textsc{Major}).

SIREN judgements are made with respect to the functional consequences of the single error rather than its linguistic quality. The presence of a HOPE error therefore does not automatically entail a loss of actionability, nor does its HOPE severity determine its SIREN impact. This separation allows us to examine which translation errors have consequences for the protective function of the emergency message and how substantial those consequences are.

The complete SIREN-span annotation procedure is illustrated in the decision tree provided in Appendix~\ref{appendix:siren-decision-tree}.

\subsubsection{Pilot Experiments on Italian--English}
\label{sec:pilot1}
To validate the protocol before full-scale annotation, we ran three pilot rounds, adopting the iterative batch-based methodology of Temnikova et al. ~\shortcite{temnikova-etal-2024-sm} in which inter-annotator agreement is computed after each batch and disagreements are reconciled before proceeding to the next round. 

In the first pilot round, two annotators independently applied the SIREN layer to 14 errors identified in the ChatGPT-4o outputs and 9 errors in the Google Translate outputs. Agreement on the binary actionability decision varied substantially across the two MT systems.
For ChatGPT-4o, the annotators agreed on 8 of 14 errors (57.14\%), yielding Cohen's $\kappa=-0.024$. For Google Translate, they agreed on 8 of 9 errors (88.89\%), with substantially higher chance-corrected agreement ($\kappa=0.769$).

Following the first pilot, a reconciliation meeting was held with both the annotators and a meta-reviewer, who did not participate in the annotation process. The disagreements were discussed qualitatively to identify recurring sources of ambiguity in the annotation protocol. Most involved cases in which an existing translation error affected lexical or terminological precision without necessarily altering the protective action conveyed by the message. Examples included uncommon lexical choices such as \emph{semi-basement} and \emph{cell phone} found in the English translations. These cases highlighted the difficulty of distinguishing general translation-quality issues from errors that have a functional impact on actionability in a real-world context. Based on the reconciliation process, the annotation guidelines were revised to make this distinction more explicit and to simplify the rationale taxonomy. In particular, the original categories \emph{Protective action distorted} and \emph{Required action unclear}, which annotators found difficult to distinguish consistently, were merged into a single category, \emph{Protective action compromised}. The revised guidelines were then applied in the second pilot round.

In the second pilot, agreement improved to 27/30 judgements (90.00\%; Cohen's $\kappa = 0.615$) for ChatGPT-4o. For Google Translate, the annotators agreed on 23/30 judgements (76.67\%), with Cohen's $\kappa = 0.516$. These results suggested that the revisions introduced after the first pilot had improved the consistency of the binary actionability judgement, particularly for ChatGPT-4o. However, the lower agreement observed for Google Translate indicated that some borderline cases remained difficult to classify consistently. The disagreements concerned non-idiomatic expressions, such as \emph{fix} instead of \emph{secure}, \emph{prudent attitude}, for which the intended protective action remained understandable despite potentially confusing wording. The second pilot therefore highlighted a difference in the decision threshold for actionability: specifically, whether additional cognitive effort or potential hesitation should be sufficient to classify an error as compromising actionability. This distinction was further clarified during reconciliation before proceeding to the third pilot. A third pilot was therefore conducted to assess whether the annotation protocol would remain stable when applied to a new batch of errors. For ChatGPT-4o, the annotators agreed on 30 of 35 judgements (85.71\%), yielding Cohen's $\kappa = 0.684$. For Google Translate, the annotators agreed on 30 of 33 judgements (90.91\%), yielding Cohen's $\kappa = 0.812$.

\subsubsection{Rollout Annotation Italian--English}

Following the three pilot rounds, the revised SIREN-span protocol was applied to the remaining Italian--English data selected for annotation. The refinements introduced during
the pilot phase clarified the distinction between general translation-quality issues and errors affecting actionability, as well as the decision threshold for borderline cases.

\paragraph{Actionability Preservation.}
In Study~1, the preservation of actionability is assessed at the error-level using SIREN-span, which asks whether an identified translation error compromises the actionability of the emergency message. Accordingly, a \textsc{Yes} judgement indicates that actionability is compromised, whereas a \textsc{No}
judgement indicates that the error does not affect actionability. The rollout comprised 196 SIREN-annotated HOPE error instances for ChatGPT-4o and 197 for Google Translate. As shown in Table~\ref{tab:siren-iten-overall}, 62 ChatGPT-4o errors (31.63\%) and 96 Google Translate errors (48.73\%) were judged to compromise actionability. Here, $N$ denotes the number of previously identified HOPE error instances evaluated with SIREN rather than the number of source sentences. The results should therefore be interpreted as the proportion of existing HOPE errors that have a functional impact on actionability, rather than as overall system-level error rates.

\begin{table}[t]
\centering
\scriptsize
\caption{SIREN-span results for the Italian--English experiment.
A YES judgement indicates that the identified translation error
compromises the actionability of the emergency message.}
\label{tab:siren-iten-overall}
\begin{tabular}{lrrrrr}
\toprule
\textbf{System} & \textbf{$N$} & \textbf{YES} & \textbf{NO} &
\textbf{YES (\%)} & \textbf{NO (\%)} \\
\midrule
ChatGPT-4o       & 196 & 62 & 134 & 31.63 & 68.37 \\
Google Translate & 197 & 96 & 101 & 48.73 & 51.27 \\
\bottomrule
\end{tabular}
\end{table}

\paragraph{Rationale Distribution.}

Among errors judged to compromise actionability, we examine the distribution of the four SIREN rationale categories. Since a \textsc{Yes} judgement in SIREN-span indicates that actionability is compromised, rationale distributions are calculated conditionally on \textsc{Yes} cases. The appropriate conditional probability is therefore:

\[ \tiny 
P(\text{category} \mid \text{YES})
=
\frac{\#\,\text{YES errors in category}}
     {\#\,\text{YES errors with a valid category}}
\times 100\%.
\]

All actionability-compromising errors received a rationale annotation, resulting in 62 categorised \textsc{Yes} cases for ChatGPT-4o and 96 for Google Translate.

\begin{table*}[t]
\centering
\caption{Distribution of SIREN rationale categories among errors judged
to compromise actionability (\textsc{Yes}) in the Italian--English experiment.}
\label{tab:siren-iten-categories}
\begin{tabular}{lrrrr}
\toprule
& \multicolumn{2}{c}{\textbf{ChatGPT-4o} ($n=62$)}
& \multicolumn{2}{c}{\textbf{Google Translate} ($n=96$)} \\
\cmidrule(lr){2-3} \cmidrule(lr){4-5}
\textbf{SIREN rationale}
& \textbf{$n$} & \textbf{\%}
& \textbf{$n$} & \textbf{\%} \\
\midrule
Directive force weakened      &  9 & 14.52 & 19 & 19.79 \\
Directive force strengthened  &  4 &  6.45 &  0 &  0.00 \\
Protective action compromised & 47 & 75.81 & 70 & 72.92 \\
Urgency marker omitted        &  2 &  3.23 &  7 &  7.29 \\
\midrule
Total                         & 62 & 100.00 & 96 & 100.00 \\
\bottomrule
\end{tabular}
\end{table*}

As shown in Table~\ref{tab:siren-iten-categories},
\emph{protective action compromised} was by far the most frequent rationale for both systems, accounting for 47 of 62 ChatGPT-4o cases (75.81\%) and 70 of 96 Google Translate cases (72.92\%). \emph{Directive force weakened} was the second most frequent rationale, observed in 14.52\% of ChatGPT-4o cases
and 19.79\% of Google Translate cases. \emph{Directive force strengthened} occurred only for ChatGPT-4o (4/62; 6.45\%), whereas \emph{urgency marker omitted} was relatively infrequent for both systems, accounting for 3.23\% of ChatGPT-4o and 7.29\% of Google Translate cases.

\paragraph{Actionability Impact.}

For errors judged to compromise actionability, we additionally examine their SIREN actionability impact. Impact is assessed independently of the original HOPE severity: an error assigned a low severity under HOPE may nevertheless have a substantial functional impact on actionability, and vice versa.
As with the rationale distribution, impact is calculated conditionally on \textsc{Yes} cases.

\begin{table}[t]
\centering
\scriptsize
\caption{Distribution of SIREN actionability-impact scores among errors
judged to compromise actionability (\textsc{Yes}) in the Italian--English
experiment. Scores 1, 2, and 3 correspond to Minor, Medium, and Major
impact, respectively.}
\label{tab:siren-iten-impact}
\begin{tabular}{lcrrrr}
\toprule
& & \multicolumn{2}{c}{\textbf{ChatGPT-4o} ($n=62$)}
& \multicolumn{2}{c}{\textbf{Google MT} ($n=96$)} \\
\cmidrule(lr){3-4} \cmidrule(lr){5-6}
\textbf{Impact} & \textbf{Score}
& \textbf{$n$} & \textbf{\%}
& \textbf{$n$} & \textbf{\%} \\
\midrule
Minor  & 1 & 38 & 61.29 & 38 & 39.58 \\
Medium & 2 & 18 & 29.03 & 39 & 40.63 \\
Major  & 3 &  6 &  9.68 & 19 & 19.79 \\
\midrule
Total & -- & 62 & 100.00 & 96 & 100.00 \\
\bottomrule
\end{tabular}
\end{table}

As shown in Table~\ref{tab:siren-iten-impact}, the two systems exhibit different actionability-impact profiles. For ChatGPT-4o, the majority of actionability-compromising errors were rated \textsc{Minor} (38/62; 61.29\%), while 18 (29.03\%) were rated \textsc{Medium} and six (9.68\%) \textsc{Major}. Google Translate showed a larger proportion of higher-impact
cases: 39 of 96 errors (40.63\%) were rated \textsc{Medium} and 19 (19.79\%) \textsc{Major}. Thus, approximately one fifth of Google Translate's actionability-compromising errors were rated \textsc{Major}, compared with approximately one tenth of ChatGPT-4o's.

\subsection{Study 2: Message-Level Actionability}
\label{sec:study2}

Study~2 evaluates actionability preservation at the level of the translated message as a whole. We use the same source material to generate new English$\rightarrow$French translations with TranslateGemma-4B, an open-weight translation model, and Google Translate. Two annotators
independently evaluate these translations using the SIREN protocol. Detailed annotator profiles are provided in Appendix~\ref{appendix:annotators}.

Unlike Study~1, which applies SIREN to individual errors already identified through HOPE, Study~2 does not rely on pre-existing error annotations. Instead, annotators compare each English source message with its French translation and assess actionability by answering the following question:
\begin{quote}
\emph{Is the actionability of the emergency message preserved in the translation?}
\end{quote}
Annotators provide a binary \textsc{Yes}/\textsc{No} judgement. A
\textsc{Yes} judgement indicates that the translation preserves the recipient's ability to identify and perform the intended protective action with the appropriate directive force and urgency, whereas a \textsc{No} judgement indicates that actionability is compromised.

\subsubsection{Prompting}
For TranslateGemma-4B, we followed the prompt format recommended in the
model's technical documentation on Ollama, adapting it to the English--French translation direction and adding a brief domain-specific instruction indicating experience with crisis-related content. No examples or additional translation instructions were provided. The complete prompt is reported in Appendix~\ref{appendix:translation-prompt}.

\subsubsection{Pilot Experiments on English--French}

Prior to the full-scale evaluation, we conducted a pilot annotation to assess the applicability of SIREN at the message level, following the iterative-batch methodology described in Section~\ref{sec:pilot1}. Two annotators
independently evaluated 13 English--French translations produced by
TranslateGemma-4B and the corresponding 13 translations produced by
Google Translate.

For TranslateGemma-4B, the annotators agreed on 10 of 13 translations
(76.92\%), yielding Cohen's $\kappa=0.316$. For Google Translate, they
agreed on 12 of 13 translations (92.31\%), with Cohen's
$\kappa=0.629$. Following the first pilot, disagreements were reviewed and the guidelines were refined accordingly. In particular, the annotators agreed that overly literal and wordy translations should be classified as instances of directive force weakening when the literal rendering reduced the directness of the instruction compared with the source message.

A second pilot was then conducted on 24 English--French translations per system. Agreement increased for both systems. For TranslateGemma-4B, the annotators agreed on 22 of 24 messages (91.67\%), with Cohen's
$\kappa = 0.48$. For Google Translate, they reached complete agreement (24/24; 100\%), with both annotators judging actionability to be preserved across all translations in the second pilot.

A third pilot was conducted on a larger batch to further assess the stability of the revised annotation protocol. For TranslateGemma-4B the annotators agreed on 47 of 51 judgements (92.16\%), yielding Cohen's $\kappa = 0.769$. For Google Translate, agreement reached 49 of 51 judgements (96.08\%), with Cohen's $\kappa = 0.811$. Taken together, the three pilot rounds show a progressive stabilisation of inter-annotator agreement for the English--French language pair, particularly for
TranslateGemma-4B, for which both observed agreement and Cohen's $\kappa$
increased across successive rounds.

\subsubsection{Rollout Annotation English--French}

Following the three pilot rounds, the annotators applied the revised SIREN-msg protocol to the full rollout English--French data. The refinements introduced during the pilot phase clarified the treatment of overly literal or wordy translations, particularly when such renderings preserved the intended protective action
but reduced the directive force of the source message.

\paragraph{Actionability Preservation.}

A total of 191 messages were evaluated for each system. As shown in Table~\ref{tab:siren-enfr-overall}, TranslateGemma-4B preserved actionability in 163 cases (85.34\%), compared with 177 cases (92.67\%) for Google Translate. Conversely, actionability was compromised in 28 TranslateGemma-4B translations (14.66\%) and 14 Google Translate translations (7.33\%). Thus, the observed rate of actionability-compromising translations was twice as high for TranslateGemma-4B:

\[
\frac{28/191}{14/191} = 2.00
\]

\begin{table}[t]
\centering
\scriptsize
\caption{SIREN-msg results for the English--French experiment.
A YES judgement indicates that the translation preserves the
actionability of the emergency message.}
\label{tab:siren-enfr-overall}
\begin{tabular}{lrrrrr}
\toprule
\textbf{System} & \textbf{$N$} & \textbf{YES} & \textbf{NO} &
\textbf{YES (\%)} & \textbf{NO (\%)} \\
\midrule
TranslateGemma-4B & 191 & 163 & 28 & 85.34 & 14.66 \\
Google Translate  & 191 & 177 & 14 & 92.67 & 7.33 \\
\bottomrule
\end{tabular}
\end{table}

\paragraph{Rationale Distribution.}

Among translations for which actionability was not preserved, we examine the distribution of the four SIREN rationale categories. Since a \textsc{No} judgement in SIREN-msg indicates that actionability is compromised, rationale distributions are calculated conditionally on \textsc{No} cases. The
appropriate conditional probability is therefore:

\[
P(\text{category} \mid \text{NO})
=
\frac{\#\,\text{NO messages in category}}
     {\#\,\text{NO messages}}
\]

\begin{table*}[t]
\centering
\caption{Distribution of SIREN categories among translations for which 
actionability was not preserved (NO). Percentages are calculated within 
the NO cases of each system.}
\label{tab:siren-enfr-categories}
\begin{tabular}{lrrrr}
\toprule
& \multicolumn{2}{c}{\textbf{TranslateGemma-4B} ($n=28$)}
& \multicolumn{2}{c}{\textbf{Google Translate} ($n=14$)} \\
\cmidrule(lr){2-3} \cmidrule(lr){4-5}
\textbf{SIREN category} 
& \textbf{$n$} & \textbf{\%} 
& \textbf{$n$} & \textbf{\%} \\
\midrule
Directive force weakened       & 23 & 82.14 & 4 & 28.57 \\
Directive force strengthened   &  0 &  0.00 & 0 &  0.00 \\
Protective action compromised  &  4 & 14.29 & 9 & 64.29 \\
Urgency marker omitted         &  1 &  3.57 & 1 &  7.14 \\
\midrule
Total                           & 28 & 100.00 & 14 & 100.00 \\
\bottomrule
\end{tabular}
\end{table*}

As shown in Table~\ref{tab:siren-enfr-categories}, the two systems exhibited different distributions of actionability-related failures. For TranslateGemma-4B, \emph{directive force
weakened} was the predominant rationale, accounting for 23 of 28
actionability-compromising translations (82.14\%). By contrast, Google Translate failures were predominantly classified as \emph{protective action compromised}, accounting for 9 of 14 cases (64.29\%). \emph{Urgency marker omitted} was infrequent for both systems, while no instances of \emph{directive force strengthened} were observed.

\paragraph{Actionability Impact.}

For translations in which actionability was not preserved, we additionally examine their SIREN actionability impact. As in Study~1, impact is classified as \textsc{Minor}, \textsc{Medium}, or \textsc{Major}, corresponding to scores of 1, 2, and 3, respectively.

\begin{table}[t]
\centering
\tiny 
\caption{Distribution of actionability-impact scores among translations for which actionability was not preserved (NO). Scores 1, 2, and 3 correspond to Minor, Medium, and Major impact, respectively.}
\label{tab:siren-enfr-impact}
\begin{tabular}{lcrrrr}
\toprule
& & \multicolumn{2}{c}{\textbf{TranslateGemma-4B} ($n=28$)}
& \multicolumn{2}{c}{\textbf{Google MT} ($n=14$)} \\
\cmidrule(lr){3-4} \cmidrule(lr){5-6}
\textbf{Impact} & \textbf{Score} 
& \textbf{$n$} & \textbf{\%}
& \textbf{$n$} & \textbf{\%} \\
\midrule
Minor  & 1 & 15 & 53.57 & 4 & 28.57 \\
Medium & 2 & 10 & 35.71 & 2 & 14.29 \\
Major  & 3 &  3 & 10.71 & 8 & 57.14 \\
\midrule
Total  & -- & 28 & 100.00 & 14 & 100.00 \\
\bottomrule
\end{tabular}
\end{table}
As shown in Table~\ref{tab:siren-enfr-impact}, most TranslateGemma-4B actionability-compromising translations were rated \textsc{Minor} (15/28; 53.57\%) or \textsc{Medium} (10/28; 35.71\%), while only three cases (10.71\%) were rated \textsc{Major}. In contrast, Google Translate exhibited a
different impact profile: 8 of its 14 actionability-compromising translations (57.14\%) were rated \textsc{Major}, compared with four \textsc{Minor} (28.57\%) and two \textsc{Medium} cases (14.29\%). Thus, although Google Translate had fewer actionability-compromising translations overall, its observed failures were more frequently associated with \textsc{Major}
functional impact.

\section{Discussion}
\label{sec:evaluation}

\textbf{Error-Level Actionability.}
The Italian--English experiment reveals differences in both the frequency and functional impact of actionability-compromising errors. Among the HOPE errors evaluated with SIREN, Google Translate errors were more frequently judged to compromise actionability than ChatGPT-4o errors (48.73\% vs.\ 31.63\%). A breakdown of actionability-compromising errors by HOPE error category is provided in Appendix~\ref{appendix:hope-type}. For both systems, these cases mainly involved the protective action itself. However, their impact profiles differed: ChatGPT-4o's actionability-compromising errors were mostly \textsc{Minor}, whereas Google Translate showed larger proportions of \textsc{Medium} and \textsc{Major} impacts. 

\paragraph{Message-Level Actionability.}
In the English--French experiment, Google Translate preserved actionability more frequently than TranslateGemma-4B (92.67\% vs.\ 85.34\%). The systems nevertheless exhibited distinct failure profiles. TranslateGemma-4B failures predominantly weakened the directive force and were primarily rated \textsc{Minor} or \textsc{Medium}. In contrast, Google Translate's failures, although less frequent, compromised more often the protective action and were more frequently assigned \textsc{Major} impact.

\paragraph{Comparisons of the Two Studies.}
In both experimental settings, SIREN identifies functional changes affecting the protective action, the directive force, and the urgency of the emergency message. While compromised protective actions were the most frequently observed rationale for both systems in the Italian--English experiment, the English--French experiment revealed different failure profiles. TranslateGemma-4B's failures mostly involved weakened directive force, whereas those of Google Translate more frequently compromised the protective action itself.

The findings show that the frequency of actionability-compromising errors or translations does not necessarily reflect their functional impact. In the English--French experiment, Google Translate preserved actionability more frequently than TranslateGemma-4B, yet a larger proportion of its compromised translations received \textsc{Major} impact ratings. More broadly, the results indicate that asking whether one MT system is simply ``better'' than another, based on aggregate scores alone, may be insufficient for high-stakes crisis communication.
Furthermore, these results demonstrate that \textit{an error identified through conventional translation-quality assessment does not necessarily constitute an actionability failure}, while also showing that \textbf{SIREN can distinguish the functional consequences of errors that do compromise the protective function of a message}.

\section{Conclusions}
This paper introduced \textsc{SIREN} (Situation-aware Index for Response and Emergency ActioNability), a human evaluation framework designed to assess whether machine-translated emergency messages preserve the action-oriented dimensions of their source. SIREN evaluates actionability preservation and, where failures are identified, characterises their functional rationale and impact.

The SIREN framework was applied in two complementary experimental settings. The Italian--English study demonstrated its applicability at error level, augmenting existing HOPE annotations with assessments of their consequences for actionability. The English--French study extended the framework to holistic message-level evaluation, independently of error annotations. The latter setting additionally produced an \emph{English--French crisis-translation resource} containing human actionability judgements and post-edited reference translations, providing a basis for further research on actionability-aware evaluation.

Together, the two studies illustrate how this framework can support the assessment of actionability at different levels of granularity. By focusing on the pragmatic consequences of translation choices, SIREN provides a complementary layer of evaluation alongside established linguistic, semantic, and automatic MT evaluation approaches, rather than replacing them or human expertise.

\section*{Limitations}
\label{sec:limitations}

This study has several limitations. First, the experiments are conducted on a relatively limited dataset, with particularly small numbers of actionability-compromising cases in some analyses. Experiments on larger datasets would provide more stable estimates and allow the robustness of the observed patterns to be assessed. Second, SIREN is evaluated on only two language pairs and a limited number of MT systems, warranting further validation across languages, crisis domains, and translation architectures. Extending the evaluation to languages beyond the Indo-European family would help assess the framework's applicability to linguistic systems in which directive force and urgency may be expressed through particles or verbal morphology. Finally, although the pilot experiments showed increasingly consistent annotation, actionability remains a human judgement; future work should adopt a reception-oriented approach with 
end-user studies to examine whether translated emergency instructions are correctly understood and acted upon.

\section*{Acknowledgments}
This work was partially supported by the PhD Programme in Humanities and Technologies funded by the University of Macerata under D.R. No 253/2023.
The second author acknowledges financial support from the HumanAI Cluster at the Leiden Institute of Advanced Computer Science (LIACS), Leiden University. The authors thank the anonymous reviewers for their feedback and suggestions, which helped improve this article.

\section*{Ethics Statement}
The authors wish to acknowledge that SIREN is a research framework currently under development and should not be interpreted as a certification of MT system safety or readiness for operational deployment. The framework is intended to support research on actionability preservation and to complement, rather than replace, human expertise and established evaluation procedures. We acknowledge that the framework and the judgments reported in this study are subject to the limitations of the data, language pairs, and human annotation process, and further validation is required before any operational application.

\bibliography{bibliography}

\clearpage
\appendix

\section{Annotator Profiles}
\label{appendix:annotators}

\subsection{Study 1: Error-Level Actionability}

Two annotators participated in the error-level SIREN annotation described in
Section~\ref{sec:study1}. Annotator~1 is a native speaker of British English with C1 proficiency in Italian and a background in modern languages and law.
Annotator~2 is a native speaker of Italian with C1 proficiency in English and a background in international cooperation.

\subsection{Study 2: Message-Level Actionability}

A separate pair of annotators participated in the message-level SIREN annotation described in Section~\ref{sec:study2}. Annotator~1 is a native speaker of French with a background in modern languages and translation.
Annotator~2 is a native speaker of Italian with C1 proficiency in French and a background in translation and crisis communication.

\section{SIREN Annotation Decision Tree}
\label{appendix:siren-decision-tree}

\begin{figure*}[!t]
    \centering
    \includegraphics[width=\textwidth]{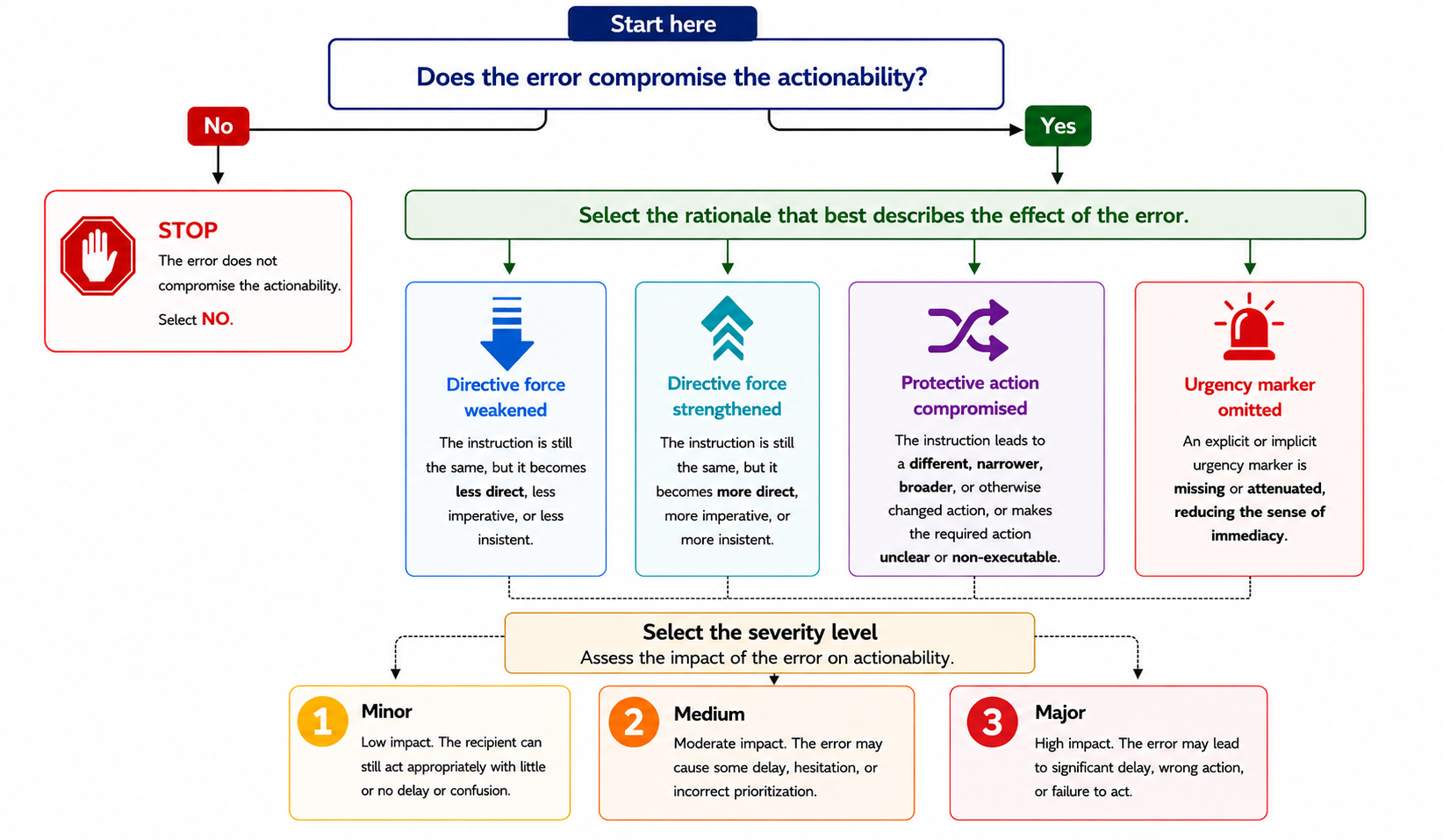}
    \caption{SIREN-span annotation decision tree. Annotators first determine
    whether an existing translation error compromises actionability. For
    actionability-compromising errors, they identify the functional rationale
    and subsequently assign an actionability impact level (\textsc{Minor},
    \textsc{Medium}, or \textsc{Major}).}
    \label{fig:siren-decision-tree}
\end{figure*}

Figure~\ref{fig:siren-decision-tree} provides the complete decision procedure
used for SIREN-span annotation.

\section{Qualitative Examples of SIREN Annotations}
\label{appendix:siren-examples}

Tables~\ref{tab:siren-qualitative-examples} and
\ref{tab:siren-enfr-examples} provide representative examples from the
Italian--English and English--French experiments, respectively, illustrating
how different translation choices can affect the functional dimensions
captured by SIREN.

\begin{table*}[t]
\centering
\small
\resizebox{\textwidth}{!}{%
\begin{tabular}{p{2.6cm}p{3.9cm}p{4.8cm}p{3.0cm}p{1.1cm}}
\toprule
\textbf{System} &
\textbf{Source} &
\textbf{MT Output} &
\textbf{SIREN Rationale} &
\textbf{Impact} \\
\midrule

ChatGPT-4o &
Non transitare lungo strade allagate: potrebbero esserci voragini, buche, tombini aperti o cavi elettrici tranciati. &
\textbf{Do not walk along} flooded streets: there may be voids, holes, open manholes, or cut electrical wires. &
Protective action compromised &
Major \\

\addlinespace

ChatGPT-4o &
Prima di uscire di casa chiudi gas, acqua e luce e indossa le scarpe. &
Before leaving your home, turn off the gas, water, and electricity, and put
on \textbf{sturdy shoes}. &
Directive force strengthened &
Medium \\

\addlinespace

Google Translate &
Usa il telefono solo per reale necessità. &
Use your phone only for \textbf{real needs}. &
Directive force weakened &
Medium \\

\addlinespace

Google Translate &
Informati sulle vie di allontanamento da seguire previste dalle
pianificazioni comunale e nazionale. &
Find out about the \textbf{distancing routes} to follow provided for by
municipal and national plans. &
Protective action compromised &
Medium \\

\bottomrule
\end{tabular}%
}
\caption{Representative SIREN-span annotations from the Italian--English
experiment. Boldface highlights translation choices relevant to the
actionability judgement.}
\label{tab:siren-qualitative-examples}
\end{table*}

\begin{table*}[t]
\centering
\small
\resizebox{\textwidth}{!}{%
\begin{tabular}{p{2.6cm}p{3.9cm}p{4.8cm}p{3.0cm}p{1.1cm}}
\toprule
\textbf{System} &
\textbf{Source} &
\textbf{MT Output} &
\textbf{SIREN Rationale} &
\textbf{Impact} \\
\midrule

TranslateGemma-4B &
Inform yourself about your municipality's Civil Protection plan. &
\textbf{Vous informer} sur le plan de protection civile de votre commune. &
Directive force weakened &
Minor \\

\addlinespace

TranslateGemma-4B &
As a general indication, indoor sheltering may last up to 48 hours. &
Pour information générale, \textbf{la recherche d'abri en intérieur}
peut durer jusqu'à 48 heures. &
Protective action compromised &
Major \\

\addlinespace

Google Translate &
If you live in the intervention area, check that your home is safe and take
all necessary steps to eliminate situations that can represent a danger in
case of an earthquake. &
Si vous habitez dans la zone d'intervention,
\textbf{vérifiez la sécurité de votre logement} et prenez toutes les
dispositions nécessaires pour éliminer les situations pouvant présenter
un danger en cas de séisme. &
Protective action compromised &
Medium \\

\addlinespace

Google Translate &
Don't use the car, it could become a trap. &
N'utilisez pas votre voiture ; elle pourrait
\textbf{se transformer en piège}. &
Directive force weakened &
Minor \\

\bottomrule
\end{tabular}%
}

\caption{Representative SIREN-msg annotations from the English--French
experiment. Examples are balanced across TranslateGemma-4B and Google
Translate. Boldface highlights translation choices relevant to the
actionability judgement.}
\label{tab:siren-enfr-examples}
\end{table*}

\section{Translation Prompt}
\label{appendix:translation-prompt}

The following prompt was used to generate the English--French machine
translations evaluated in Study~2. The source emergency message was
inserted in the final placeholder.

\begin{tcolorbox}[
    breakable,
    colback=white,
    colframe=gray!70,
    boxrule=1pt,
    arc=3mm,
    left=3mm,
    right=3mm,
    top=3mm,
    bottom=3mm
]
\ttfamily
You are a professional English (en) to French (fr) translator with
experience translating crisis-related content. 

Your goal is to accurately convey the meaning and nuances of the original English text while adhering to French grammar, vocabulary, and cultural sensitivities.

Produce only the French translation, without any additional explanations
or commentary.

Please translate the following English text into French:

\{source\_text\}
\end{tcolorbox}

\section{Pilot Study IAA Scores}
\label{appendix:pilot-iaa}

Table~\ref{tab:pilot-iaa} reports inter-annotator agreement (IAA) scores
across the three pilot annotation batches for both the Italian--English
and English--French experiments.

\begin{table*}[!t]
\centering
\small

\begin{minipage}[t]{0.48\textwidth}
\centering
\begin{tabular}{clccc}
\toprule
\textbf{Batch} & \textbf{System} & \textbf{$N$} &
\textbf{Agreement} & \textbf{$\kappa$} \\
\midrule
\multirow{2}{*}{1}
 & ChatGPT-4o       & 14 & 57.14\% & $-0.024$ \\
 & Google Translate &  9 & 88.89\% & 0.769 \\
\cmidrule(lr){1-5}
\multirow{2}{*}{2}
 & ChatGPT-4o       & 30 & 90.00\% & 0.615 \\
 & Google Translate & 30 & 76.67\% & 0.516 \\
\cmidrule(lr){1-5}
\multirow{2}{*}{3}
 & ChatGPT-4o       & 35 & 85.71\% & 0.684 \\
 & Google Translate & 33 & 90.91\% & 0.812 \\
\bottomrule
\end{tabular}

\vspace{2mm}

\small
\textbf{Italian--English.}
$N$ denotes the number of HOPE error instances evaluated with SIREN.
\end{minipage}
\hfill
\begin{minipage}[t]{0.48\textwidth}
\centering
\begin{tabular}{clccc}
\toprule
\textbf{Batch} & \textbf{System} & \textbf{$N$} &
\textbf{Agreement} & \textbf{$\kappa$} \\
\midrule
\multirow{2}{*}{1}
 & TranslateGemma-4B & 13 & 76.92\% & 0.316 \\
 & Google Translate  & 13 & 92.31\% & 0.629 \\
\cmidrule(lr){1-5}
\multirow{2}{*}{2}
 & TranslateGemma-4B & 24 & 91.67\% & 0.480 \\
 & Google Translate  & 24 & 100.00\% & -- \\
\cmidrule(lr){1-5}
\multirow{2}{*}{3}
 & TranslateGemma-4B & 51 & 92.16\% & 0.769 \\
 & Google Translate  & 51 & 96.08\% & 0.811 \\
\bottomrule
\end{tabular}

\vspace{2mm}

\small
\textbf{English--French.}
$N$ denotes the number of messages evaluated with SIREN.
\end{minipage}

\caption{Inter-annotator agreement across the three pilot annotation batches
for the Italian--English and English--French experiments.}
\label{tab:pilot-iaa}
\end{table*}

\section{Actionability Loss by HOPE Error Type}
\label{appendix:hope-type}

Table~\ref{tab:hope-type-yes} reports, for each system in the Italian--English rollout, how many errors for each HOPE type were judged to compromise actionability (\textsc{Yes}).

\begin{table*}[t]
\centering
\small
\setlength{\tabcolsep}{14pt}
\renewcommand{\arraystretch}{1.15}
\begin{tabular}{llrrrr}
\toprule
& & \multicolumn{2}{c}{\textbf{ChatGPT-4o}}
& \multicolumn{2}{c}{\textbf{Google Translate}} \\
\cmidrule(lr){3-4} \cmidrule(lr){5-6}
\textbf{Code} & \textbf{HOPE error type}
& $N$ & \textsc{Yes} (\%)
& $N$ & \textsc{Yes} (\%) \\
\midrule
IMP & Impact         & 16 & 5 (31.25)   & 18 & 11 (61.11) \\
TRM & Terminology    & 40 & 12 (30.00)  & 33 & 17 (51.52) \\
UGR & Ungrammatical  & 4  & 1 (25.00)   & 12 & 5 (41.67)  \\
MIS & Mistranslation & 23 & 9 (39.13)   & 29 & 14 (48.28) \\
STL & Style          & 84 & 25 (29.76)  & 87 & 38 (43.68) \\
PRF & Proofreading   & 29 & 10 (34.48)  & 18 & 11 (61.11) \\
PRN & Proper Name    & 0  & --          & 0  & --         \\
\midrule
\multicolumn{2}{l}{\textbf{Total}}
& \textbf{196} & \textbf{62 (31.63)}
& \textbf{197} & \textbf{96 (48.73)} \\
\bottomrule
\end{tabular}
\caption{Actionability-compromising errors by HOPE error type in the Italian--English rollout. $N$ indicates the total number of errors in each category. \textsc{Yes} indicates the number of errors judged to compromise actionability, with the percentage relative to $N$ in parentheses.}
\label{tab:hope-type-yes}
\end{table*}

\end{document}